\documentclass[conference]{IEEEtran}

\usepackage{cite}
\usepackage{amsmath,amssymb,amsfonts}
\usepackage{soul}
\usepackage{algorithmic}
\usepackage{graphicx}
\usepackage{textcomp}
\usepackage{xcolor}
\usepackage{ulem}
\usepackage{enumitem}
\usepackage{url}
\usepackage{balance}

\renewcommand{\baselinestretch}{1.0}
\def\BibTeX{{\rm B\kern-.05em{\sc i\kern-.025em b}\kern-.08em
    T\kern-.1667em\lower.7ex\hbox{E}\kern-.125emX}}

\title{\texttt{DeltaNN}: Assessing the Impact of Computational Environment Parameters on the Performance of Image Recognition Models\thanks{Authors, Ajitha Rajan and Nikolaos Louloudakis, would like to acknowledge the support received from funding sources, UKRI Trustworthy Autonomous Systems Node in Governance and Regulation (EP/V026607/1) and Royal Society Industry Fellowship, for this work.}}

\author{\IEEEauthorblockN{Nikolaos Louloudakis}
\IEEEauthorblockA{\textit{n.louloudakis@ed.ac.uk} \\
\textit{University of Edinburgh\vspace{-3.5\baselineskip}}} \\
\and
\IEEEauthorblockN{Perry Gibson}
\IEEEauthorblockA{\textit{perry.gibson@glasgow.ac.uk} \\
\textit{University of Glasgow\vspace{-3.5\baselineskip}}} \\
\and
\IEEEauthorblockN{Jos\'e Cano}
\IEEEauthorblockA{\textit{jose.canoreyes@glasgow.ac.uk} \\
\textit{University of Glasgow\vspace{-3.5\baselineskip}}} \\
\and
\IEEEauthorblockN{Ajitha Rajan}
\IEEEauthorblockA{\textit{arajan@ed.ac.uk} \\
\textit{University of Edinburgh\vspace{-3.5\baselineskip}}} \\
}

\IEEEoverridecommandlockouts
\begin{document}
\bstctlcite{IEEEexample:BSTcontrol}

\maketitle


\begin{abstract}

Image recognition tasks typically use deep learning and require enormous processing power, thus relying on hardware accelerators like GPUs and TPUs for fast, timely processing. 
Failure in real-time image recognition tasks can occur due to sub-optimal mapping on hardware accelerators during model deployment, which may lead to timing uncertainty and erroneous behavior.
Mapping on hardware accelerators is done using multiple software components like deep learning frameworks, compilers, and device libraries, that we refer to as the computational environment. 
Owing to the increased use of image recognition tasks in safety-critical applications like autonomous driving and medical imaging, it is imperative to assess their robustness to changes in the computational environment, as the impact of parameters like deep learning frameworks, compiler optimizations, and hardware devices on model performance and correctness is not yet well understood.

In this paper we present a differential testing framework, \texttt{DeltaNN}, that allows us to assess the impact of different computational environment parameters on the performance of image recognition models during deployment, post training. 
\texttt{DeltaNN} generates different implementations of a given image recognition model for variations in environment parameters, namely, deep learning frameworks, compiler optimizations and hardware devices and analyzes differences in model performance as a result.
Using \texttt{DeltaNN}, we conduct an empirical study of robustness analysis of three popular image recognition models using the ImageNet dataset. 
We report the impact in terms of misclassifications and inference time differences across different settings. 
In total, we observed up to 100\% output label differences across deep learning frameworks, and up to 81\% unexpected performance degradation in terms of inference time, when applying compiler optimizations.

\end{abstract}


\section{Introduction}

\begin{figure}[t]
 \centering
 \includegraphics[width=0.9\columnwidth]{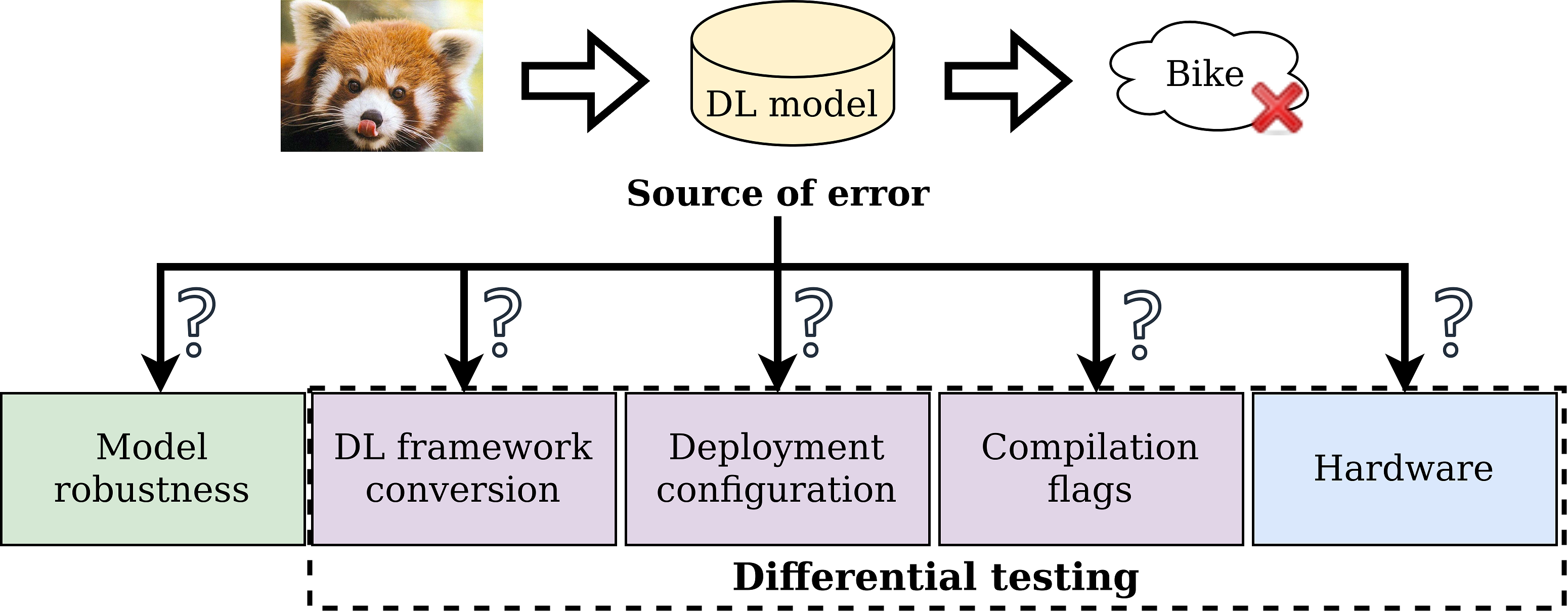}
 \caption{\label{fig:motivation} Possible sources of errors when deploying DNN models.}
 \vspace{-15pt}
\end{figure}

Much of the existing literature for assessing robustness and safety of image recognition models has focused on testing the Deep Neural Network (DNN) structure and addressing biases in the training dataset through adversarial examples and data augmentation~\cite{zhang2018deeproad, tian2018deeptest, guoCoverageGuidedDifferential2021}.
However, the impact of computational environment aspects related to the DNN model deployment process, post training, has not yet been explored. 
In particular, existing techniques fail to consider model output errors that could potentially be caused by interactions of the DNN model with the underlying computational environment -- conversions between Deep Learning (DL) frameworks (e.g., TensorFlow, PyTorch, TensorFlow Lite), compiler optimizations (e.g., operator fusion, loop unrolling, etc.), and the hardware platforms they run on (e.g., CPUs, GPUs, etc.).
Figure~\ref{fig:motivation} shows potential sources of error in the computational environment when a DNN model is deployed. 
These environment aspects are important considerations in model maintenance and evolution.

\begin{figure*}
 \centering
 \includegraphics[width=0.95\textwidth]{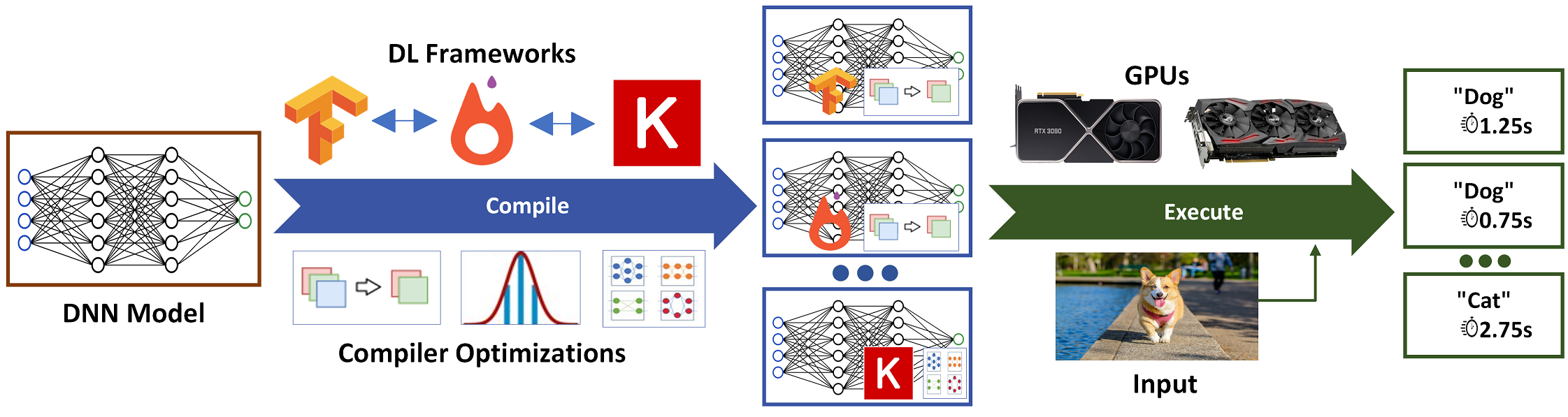}
 \caption{\label{fig:differential_testing} Differential Testing applied by \texttt{DeltaNN} for a DNN model, varying  deep learning frameworks, compiler optimizations, and hardware devices.}
\end{figure*}

To understand the impact of the computational environment on model deployment, we present a differential testing framework, \texttt{DeltaNN}\footnote{The source code is available at: \url{https://github.com/luludak/DeltaNN}}, that helps evaluate the robustness of image recognition models to changes in specific aspects of the computational environment (Figure~\ref{fig:differential_testing}).
\texttt{DeltaNN} takes as input a trained DNN model defined in a given DL framework and produces different implementations by changing the following parameters in the computational environment: 

\begin{description}[wide=0\parindent]
    \item[DL frameworks:] transforming a model defined in one DL framework to the model format of another framework. 
    Studying the impact of model conversion between DL frameworks is important, as developers may often convert their models to support resource constrained environments on mobile and IoT devices. 
    Automated conversion processes suffer from faults, mainly caused by unsupported operations in the target framework, or by the converter.
    We generate different implementations of every trained model with several popular DL frameworks. 
    \item[Compiler optimizations:] considering different levels of compiler optimizations with each generating a distinct code implementation. 
    The focus of our experiments is on graph-level optimizations like operator fusion, eliminating common subexpressions, data layout transformations for better memory utilization and access patterns on target devices, or potentially unsafe optimizations such as ``fast-math''.
    Compiler optimizations are expected to improve model performance, sometimes at the cost of model accuracy.
    We study this parameter to understand the extent of impact on model performance and correctness.
    \item[Hardware devices:] we generate implementations for a range of GPU accelerators, from a resource constrained mobile GPU to a powerful server-class GPU~\cite{yaneva2017compiler}. 
    We consider different types of devices to check if GPU specifications can impact model output.
\end{description}

We assess the robustness of the DNN models with respect to consistency in output labels for changes in each of the three computational environment parameters.
Note that it is important to check changes in the output label, as it directly affects model accuracy. 
Additionally, we monitor model inference times for different settings of computational environment parameters to understand the extent of variation among them.
Inference times are an important consideration for timing safety in real-time perception systems in applications like self-driving cars, where there is a performance requirement for object detection models to return results within a fixed time budget~\cite{dreossi2019verifai}.

We assess the robustness of three widely used image recognition models -- MobileNetV2~\cite{mobilenetv2}, ResNet101V2~\cite{resnetv2}, and InceptionV3~\cite{inceptionv3}, on the ImageNet object detection test dataset (ILSVRC2017)~\cite{ILSVRC17}.
We chose the three models based on their popularity but also to provide variety on layer architecture and model size. The dataset we selected is a competition test dataset designed to extensively test models.
The \texttt{DeltaNN} framework uses the Apache TVM~\cite{tvm} machine learning compiler stack, as it allows importing models from all major DL frameworks while providing fine-grained control over compilation configurations and execution of the DNN models, as well as a wide range of hardware backend support.

Overall, we find that conversions between DL frameworks significantly impacts output labels of the DNN models by up to 100\%. 
We identify the source of the label-impacting error  -- small amounts of noise introduced in the weights by the framework conversion tools.
The weight differences, although small, may be caused by floating-point rounding errors which can cause label changes when accumulated across the layers.
On the other hand, we found that varying hardware accelerators and compiler optimizations do not affect model output but can lead to a non-negligible performance degradation with respect to inference time under specific scenarios. 
We observed up to 81\% unexpected performance degradation in model inference times when applying certain compiler optimizations. 

\noindent In summary, we make the following contributions: 
\begin{enumerate}
    \item Assess robustness of image recognition model \textit{outputs}, post training, with respect to changes in the computational environment: DL frameworks, compiler optimizations, and hardware devices using a differential testing framework, \texttt{DeltaNN}. 
 
    \item Assess robustness of model \textit{inference time} with respect to changes in the computational environment: DL frameworks, compiler optimizations, and hardware devices.
    
    \item Analyze and identify sources of \textit{label discrepancy} when converting between DL frameworks. 
\end{enumerate}

\section{Background}

Figure~\ref{fig:systems_stack} gives an overview of the typical layers in the deep learning systems stack~\cite{iiswc_2018}.
Much of the existing work on DL model robustness has focused on testing robustness with respect to the top two layers, \textit{Datasets} and \textit{Models}. 
In this paper, we consider robustness with respect to the bottom three layers which make up the computational environment required for executing a given DNN model, which includes the deep learning framework, the related systems software, and the underlying hardware.


\subsection{Deep Learning Frameworks}

Deep Learning Frameworks (the third layer in Figure~\ref{fig:systems_stack}), provide utilities such as model declaration, training, and inference to machine learning engineers.
For our study, we consider four widely used DL frameworks: 
\textit{Keras}, \textit{PyTorch}, \textit{TensorFlow (TF)}, and \textit{TensorFlow Lite} (\textit{TFLite}).

\textbf{Keras}~\cite{chollet2015keras} is a high-level DL framework that provides APIs for effective deep learning usage. 
It acts as an interface for TensorFlow, and we aim to observe potential overheads and bug introductions from the extra layer of complexity.

\textbf{PyTorch}~\cite{pytorch} is an open source machine learning framework based on the Torch library.
It supports hardware acceleration for tensor computing operations.

\textbf{TensorFlow (TF)}~\cite{tensorflow2015-whitepaper} is an open-source DL framework developed by Google, widely used for training and inference of DNNs.

\textbf{TensorFlow Lite (TFLite)}~\cite{tensorflow2015-whitepaper} is a lightweight version of TF, and part of the full TF library, focused only on the inference of DNNs on mobile and lightweight devices.

\begin{figure}[!t]
 \centering
 \includegraphics[width=0.9\columnwidth]{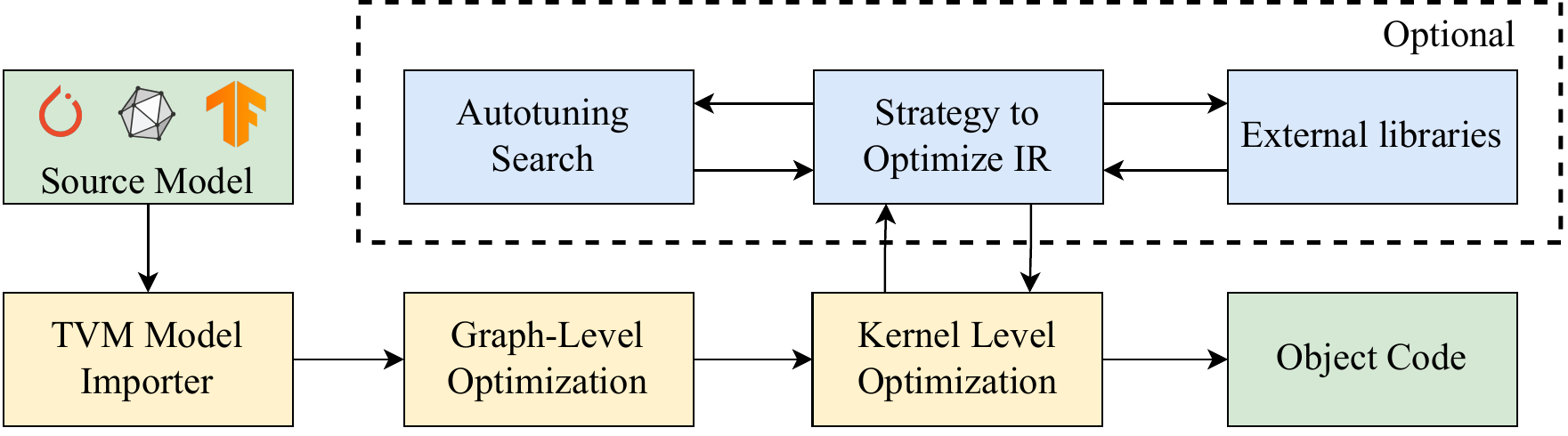}
 \caption{\label{fig:tvm_simple} Overview of DNN compilation in Apache TVM.}
\end{figure}


\subsection{Framework Conversions}

Developers and researchers convert DNN models from one DL framework to another for one of the following two reasons, 
1) portability,  to enable the compilation and execution of models on devices of varying capabilities, and
2) differing framework capabilities.
For example, some DL frameworks may provide  detailed debugging and profiling capabilities for development while others may be better suited for optimized deployment.

Conversion of DNN models between DL frameworks is facilitated by automated conversion processes enabled by tools such as \texttt{tf2onnx}~\cite{tf2onnx}, \texttt{onnx2keras}~\cite{onnx2keras}, \texttt{onnx2torch}~\cite{onnx2torch}, and \texttt{MMdnn}~\cite{liuEnhancingInteroperabilityDeep2020}. 
The conversion process can, however, suffer from errors in the model parameters and graph representation which can potentially affect the output labels.
We refer to models defined within a given DL framework as a ``\textit{native model}'', and models that have been converted to another DL framework as a ``\textit{converted model}''.
Systems such as ONNX~\cite{onnxsite} and \textit{MMdnn}~\cite{liuEnhancingInteroperabilityDeep2020} attempt to provide common intermediate formats for conversion between DL frameworks.
However, these systems can still be error-prone and have issues supporting bespoke operators, motivating our investigation of framework conversion errors.


\subsection{Systems Software: Apache TVM}

Apache TVM~\cite{tvm} is an end-to-end machine learning compiler framework for CPUs, GPUs, and specialized accelerators, actively used and supported by a wide community of developers and researchers.
It generates optimized code for specific DNN models and hardware backends, allows us to import DNN models from a range of DL frameworks, and provides profiling utilities such as per-layer inference times.
A simplified representation of Apache TVM can be seen in Figure~\ref{fig:tvm_simple}.
TVM's support of several DL frameworks, optimization settings, and hardware accelerators makes it a suitable choice to leverage within \texttt{DeltaNN}.
It also provides direct importers for models from most popular DL frameworks, which load the models as a TVM computation graph that can be optimized and compiled.
TVM also provides a set of graph-based optimizations for DNNs, such as operator fusion, elimination of common subexpressions, loop re-ordering and unrolling, tiling, vectorization, and potentially unsafe precision optimizations such as fast math.
TVM applies such optimizations automatically via the use of an \texttt{-o[0-4]} flag.
Finally, TVM supports third-party operator kernel libraries such as cuDNN~\cite{chetlur2014cudnn} and the Arm Compute Library~\cite{ComputeLibrary2022}.


\subsection{The Perception AI Models}
\label{subsec:backg_models}

A common benchmark for Perception AI models is the ImageNet image classification dataset~\cite{ILSVRC17}, which requires assigning one of 1000 possible class labels to RGB images of $224\times224$ pixels.
For solving Perception AI problems, such as classification and semantic segmentation, convolutional neural networks (CNNs) are commonly used, which are DNNs characterized by convolutional layers.
Transformer-based architectures~\cite{vaswani2017attention} have begun to provide competitive results in recent years~\cite{daiCoAtNetMarryingConvolution2021,zhaiScalingVisionTransformers2022}, however are still maturing.
Thus, for our evaluation we explore three widely used CNN models: MobileNetV2~\cite{mobilenetv2}, ResNet101V2~\cite{resnet}, and InceptionV3~\cite{inceptionv3}.
These models are widely known and extensively used for classification and semantic segmentation operations, and are the ``backbone network'' for other tasks such as object detection~\cite{chiuMobilenetSSDv2ImprovedObject2020}. 
All three models have native definitions within the DL frameworks under study.
The accuracy of the native version of each model is shown in Table~\ref{tab:native_accuracy}.
It is expected that the same model may have varying accuracy between frameworks, as each framework will define and train their own version of the model from scratch, which produce different parameters (since training is stochastic), and there may even be small differences in the graph definition (e.g., different padding parameters).
We observe that TF and TFLite models have the same accuracy, suggesting that the latter models were converted from the former by developers.

\begin{table}[]
\caption{Inference accuracy of native models on the ImageNet dataset.}
\centering
\fontsize{9}{11}\selectfont
\begin{tabular}{|l|c|c|c|c|}
\hline
\rotatebox[origin=c]{0}{{\textbf{DNN Model}} / {\textbf{Framework}}} & \rotatebox[origin=c]{90}{~PyTorch~} & \rotatebox[origin=c]{90}{Keras} & \rotatebox[origin=c]{90}{TF} & \rotatebox[origin=c]{90}{TFLite} \\\hline
   ResNet101 & 81.9 & 76.4 & 77.0 & 77.0  \\\hline
   InceptionV3 & 77.3 & 77.9 & 78.0 & 78.0  \\\hline
   MobileNetV2 & 72.2 & 71.3 & 71.9 & 71.9  \\\hline
\end{tabular}
\bigskip
\label{tab:native_accuracy}
\end{table}

\section{Related Work}

\begin{figure}[!t]
 \centering
 \includegraphics[width=0.99\columnwidth]{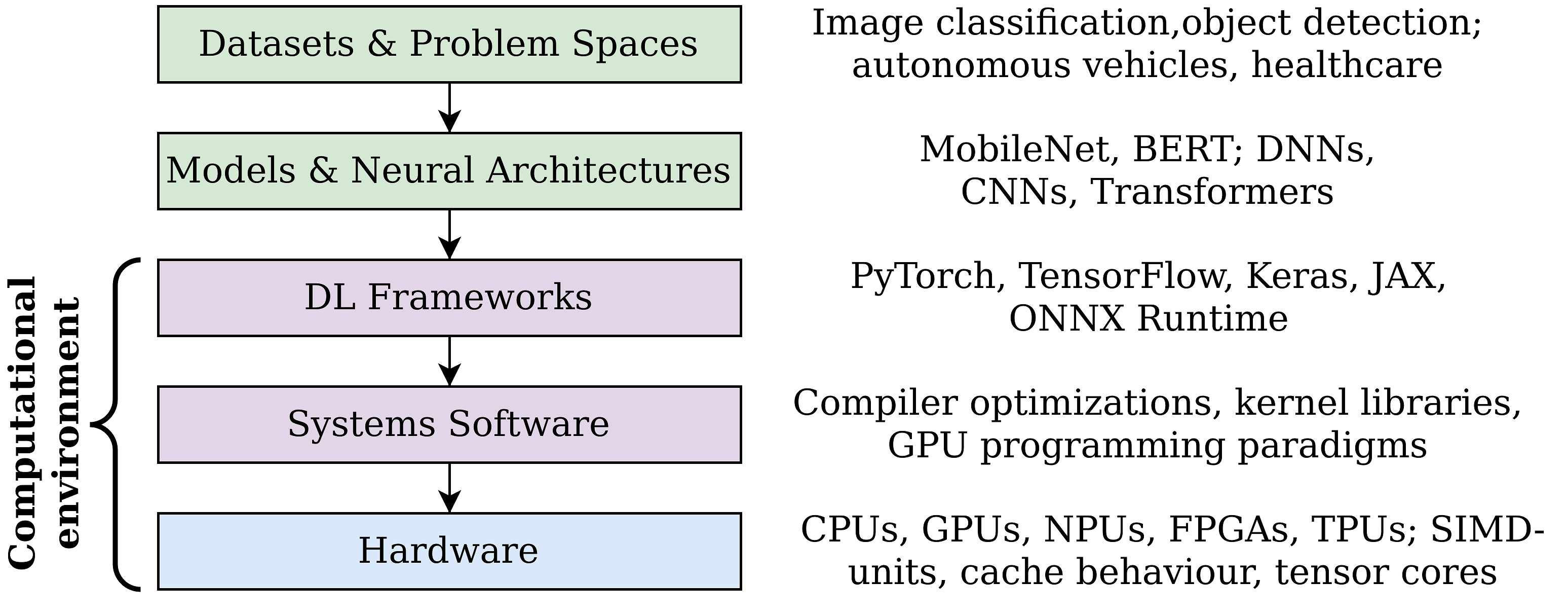}
 \caption{\label{fig:systems_stack} Relevant layers in the deep learning systems stack~\cite{iiswc_2018}.}
\end{figure}

Existing work has primarily focused on the robustness of the dataset and model architecture layers, the top two layers in Figure~\ref{fig:systems_stack}.
In particular, a survey by Zhang et al.~\cite{zhang2019machine} comprehensively presents existing testing techniques in machine learning by exploring a number of contributions in terms of correctness, robustness, and fairness, primarily focusing on model training and validation datasets.
DeepXPlore~\cite{deepxplore} applies whitebox testing by measuring neuron coverage, identifying similar DNNs for cross-reference and generating adversarial inputs to detect faults.
DLFuzz~\cite{guoCoverageGuidedDifferential2021} attempts to minutely mutate inputs to improve neuron coverage.
DeepTest~\cite{tian2018deeptest} modifies images using linear \& affine transformations, and generates inputs simulating different weather conditions and phenomena to stress-test DNNs utilized for autonomous driving. 
DeepRoad~\cite{zhang2018deeproad} applies in the same context, while using GAN-based metamorphic testing that simulate extreme weather conditions, such as heavy rain and snow.
Ayaz et al.\cite{ayaz2023ijcnn} propose to improve the robustness against adversarial attacks with deeply quantized DNNs. 
For a more comprehensive overview of adversarial inputs of DNNs, we refer the readers to a survey~\cite{shorten2019survey}.

In comparison, robustness with respect to the computational environment (last three layers of Figure~\ref{fig:systems_stack}) has received little attention. 
With respect to the DL Frameworks layer, some research has been conducted in the direction of analysis of model training and inference performance~\cite{benchmarkdlswtools,benchmarkingdlframeworkmetrics,collie2020m3, stratis2018speeding, dlbench}. 
In addition, a recent survey~\cite{studydlframeworks} explores various parameters and their effect towards model accuracy and execution time.
In terms of automated testing of DL frameworks, there are some works aiming to detect and localize inconsistencies between models sourced from different DL frameworks.
Under this context, CRADLE~\cite{pham2019cradle} applies output and model execution comparison, while LEMON~\cite{wang2020lemon} utilizes mutation testing to detect bugs in DL frameworks. 
Similarly, Audee~\cite{audee} aims to detect logical, not-a-number bugs, and crashes by applying an exploratory approach in combination with mutation testing.
However, all these contributions conduct experiments that do not consider DL framework conversions, and focus on a specific set of bugs instead of broadly exploring potential issues related to DL frameworks.
They also do not consider the impact of other computational environment aspects, such as optimizations and deployment on different hardware acceleration devices - aspects explored in our contribution.


In terms of DL Framework conversions, there is a variety of tools available in the community for that purpose.
\textit{MMdnn}\cite{liuEnhancingInteroperabilityDeep2020} is a tool focusing on the process of library conversions, using an intermediate representation. 
There are many other tools for DNN framework conversions such as \texttt{tf2onnx}\cite{tf2onnx}, \texttt{onnx2keras}\cite{onnx2keras}, \texttt{onnx2torch}\cite{onnx2torch}, as well as native APIs of TFLite and PyTorch. 
However, the error proneness of the process is overlooked in the literature, as there is only one empirical study of DL framework conversions~\cite{dlfconversionsstudy} which focuses only on conversions between ONNX and CoreML, finding prediction accuracy of converted models to be similar to the original ones. 
We explore the effects of DL framework conversions extensively in our work.

Regarding the systems software layer in Figure~\ref{fig:systems_stack}, a recent study~\cite{dlcompileroptimisationbugs} examined bugs introduced by different DL compilers.
Incorrect optimization code logic accounted for 9\% of the bugs introduced by compilers. 
Other compiler bugs presented in the study include misconfiguration, type problem, API misuse, incorrect exception handling, and incompatibility.
In our contribution, we primarily examine the effect of changing compiler optimizations on model performance, in terms of accuracy and inference time.

Finally, for the hardware layer in Figure~\ref{fig:systems_stack}, a taxonomy of faults encountered in DNNs used in object detection has been established~\cite{humbatova2020taxonomy}. The authors surveyed commits, issues and pull requests from 564 GitHub projects and 9,935 posts from Stack Overflow and interviewed 20 researchers and practitioners. 
The study revealed \textit{GPU
related bugs} to be one of the five main categories faults in deep learning tasks like object detection.
The study, however, did not explore the impact of these bugs on model performance. 
The other four categories of faults were \textit{API}, \textit{Model}, \textit{Tensors and Inputs}, and \textit{Training} which relate to the top two layers in Figure~\ref{fig:systems_stack}.

We explore in-depth the effect of DL framework conversions in terms of output label predictions and execution times under different configurations of DNN models, DL frameworks, and hardware acceleration devices capabilities during model deployment.
This extends our previous work~\cite{louloudakis2022exploring, louloudakisassessing, louloudakis2023fault} with further experiments and exploration of library conversions in detail.
To the best of our knowledge, this is the first work that assesses the effects of the computational environment aspects on image recognition models, post training.

\section{Methodology}
\label{methodology}
\begin{figure*}[!t]
 \centering
 \includegraphics[width=0.95\linewidth]{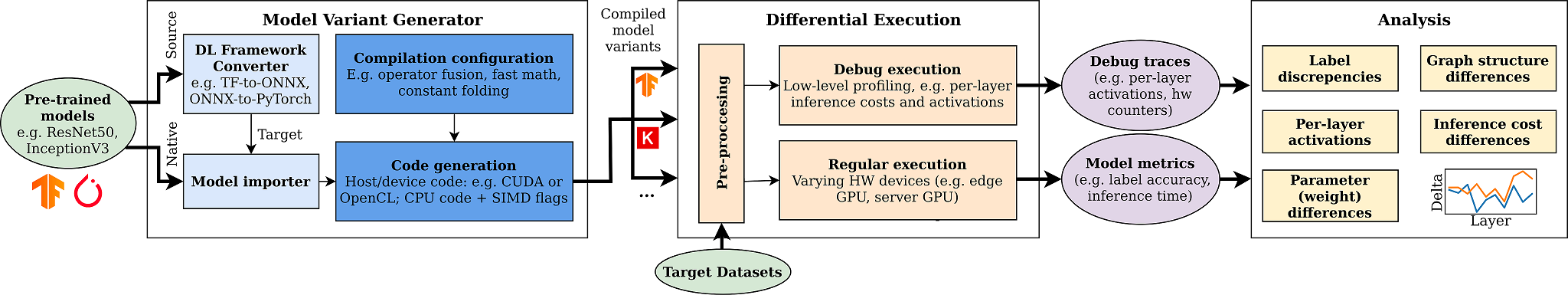}
 \caption{
 Architecture of the \texttt{DeltaNN} framework: 
 (1) \textit{Model Variant Generator} for generating different model implementations when changing and converting DL frameworks and compiler optimizations; 
 (2) \textit{Differential Execution} for executing the various model implementations on images from a target dataset; and 
 (3) \textit{Analysis} for comparing output labels and inference time between executions while analyzing source of discrepancy.
 }
 \label{fig:delta_dnn_v1}
\end{figure*}

\texttt{DeltaNN} comprises three stages, as shown in Figure~\ref{fig:delta_dnn_v1}:
(1) \textit{Model Variant Generator} that generates different equivalent model implementations when changing DL frameworks and compiler optimizations;
(2) \textit{Differential Execution} that executes each of the model implementations with images from a test dataset; and
(3) \textit{Analysis} that compares the output labels, inference time, and other data from the different implementations, and aids in localization of discrepancy sources, if any. 

\subsection{Model Variant Generator}

As seen on the left-hand side of Figure~\ref{fig:delta_dnn_v1}, the \textbf{Model Variant Generator} takes as input a pre-trained image recognition model (e.g., InceptionV3) sourced from a given DL framework (e.g., PyTorch). 
If we use one of these pre-trained models ``as-is'', and pass it directly to the \textbf{Model Importer}, we refer to it as a \textit{native} model.
However, if we convert it using the \textbf{DL Framework converter}, we refer to the original model as the \textit{source}, and the converted model as the \textit{target}.
For example, we could convert an InceptionV3 model sourced from PyTorch to the TensorFlow model format.
Across the four DL frameworks we support in \texttt{DeltaNN}, we have conversion paths from every framework to every other one.

To implement the conversions, we use the popular ONNX format~\cite{onnxsite} as an intermediate representation when a direct conversion is not possible. 
Some DL frameworks, such as PyTorch and TFLite, have native tools for this conversion; whereas for others, such as TensorFlow, we leverage popular third-party conversion tools like \texttt{tf2onnx}~\cite{tf2onnx}.
We then convert from ONNX to the target DL framework model format using a number of widely used libraries, such as \texttt{onnx2torch}~\cite{onnx2torch}, \texttt{onnx2keras}~\cite{onnx2keras}.

\paragraph{Compiler Configuration}
\texttt{DeltaNN} generates model implementations with different levels of TVM graph-level compiler optimizations: basic, default, and extended variants.
    \textbf{Basic} (\texttt{o0}) applies only ``inference simplification'', which generates simplified expressions with the same semantic equivalence as the original DNN.
    \textbf{Default} (\texttt{o2}) applies all optimizations of \texttt{o0}, as well as operator fusion for operations such as ReLU activation functions, as well as constant and scale axis folding.
    The optimizations are applied to TVM's Relay intermediate representation (IR)~\cite{tvmrelay}.
    \textbf{Extended} optimization (\texttt{o4}) applies all optimizations from Default, as well as additional ones such as eliminating common subexpressions, applying canonicalization of operations, combining parallel convolutions, dense matrix and batch matrix multiplication operations, and enabling ``fast math'' (which allows the compiler to break strict IEEE standard~\cite{ieee754} compliance for float operations if it could improve performance).
We can also enable and disable specific optimizations at a fine-grained level, which can be useful for localization.
In addition, kernel-level optimizations such as schedules, auto-tuning~\cite{chenLearningOptimizeTensor2018}, third-party libraries (such as cuDNN~\cite{chetlur2014cudnn}), and auto-scheduling~\cite{zhengAnsorGeneratingHighPerformance2020,gibsonTransferTuningReusingAutoSchedules2022} can be explored, but are not the focus of this study.

\paragraph{Code Generation}
The final part of the model variant generator takes the selected compiler configuration and imported model format and generates both host and device code, with the option to explore different programming paradigms (e.g., CUDA and OpenCL), CPU-side optimization flags (e.g., enabling vector instructions), and hardware devices (e.g., different GPU devices). 
The code generation step produces the outputs of the whole \textbf{Model Variant Generator} stage, namely several model variants, each with a different setting for compiler optimization, DNN model source or target, and host/device code configuration.


\subsection{Differential Execution}

The next stage of \texttt{DeltaNN} is \textbf{Differential Execution} of the model variants from the previous stage. 
It consists of three main steps:
(1) the \textbf{Pre-processing} module responsible for normalizing inputs for better model performance (with a variety of pre-processing functions to choose from);
(2) the \textbf{Regular Execution} module that executes the model for different target devices; and 
(3) the \textbf{Debug Execution} module, that executes the model similar to the \textit{Regular Execution} module but additionally generates execution profiling information that can be used for deeper performance and error insights.

\paragraph{Pre-processing}
It is a common practice to pre-process the inputs from the dataset before inference, similar to training. 
Examples of pre-processing include image resizing, input image pixels normalization, and more.
By default, the module pre-processes the inputs based on the model architecture and source DL framework.

\paragraph{Regular Execution}
This module executes the model on a specified target device to perform inference against a specific input and generate an output prediction. 
Execution encompasses model loading, setting up execution parameters from configuration, and experiment management (i.e., multiple runs). 
The output from this module is \textbf{Model metrics}, namely label accuracies and inference times for each image executed on the model.
This module orchestrates and performs model execution in bulk, executing inference of a whole dataset against the numerous module variants generated by the \textbf{Model Variant Generator} module.

\paragraph{Debug Execution}
As the main purpose of \texttt{DeltaNN} is differential testing, generating execution-based metadata is vital for analyzing possible sources of error in the model variants. 
The \textbf{Debug Execution} module performs model execution similar to the \textbf{Regular Execution} module.
Nevertheless, during execution, the module generates profiling metadata and debug metrics associated with the inference process, such as tensor outputs of each layer, per-layer inference time, and hardware counters.
This information is passed on as \textit{debug traces} to the \textbf{Analysis} stage for fault localization. 


\subsection{Analysis}

For every pair of model variants, the \textbf{Analysis} stage compares labels and inference time from \textit{Model metrics} for all images in the dataset. 
To compare labels, we compare the top ranked predictions between the model variants or performing rank-biased overlap to compare rankings for top-K elements. 
This means not only can we detect divergence, but also measure the level of divergence.
When an image generates different labels or inference time between a pair of model variants, the \textbf{Analysis} module compares the \textit{debug traces} from the model variants inspecting differences in \textit{per-layer activations}, \textit{weights}, and the \textit{graph structure}. 
For \textit{per-layer activations}, we compare mean, max, and standard deviations statistics of the layer activations between the pairs of models. 
The \textbf{Analysis} module also provides the capability to visualize the differences observed in layer activations and weights.

\section{Experiments}
\label{sec:experiments}
We consider three widely used image recognition models of various sizes: MobileNetV2~\cite{mobilenetv2}, ResNet101V2~\cite{resnet, resnetv2}, and InceptionV3~\cite{inceptionv3}.
We use models pre-trained on ImageNet~\cite{deng2009imagenet}, using native model definitions and pre-trained parameters/weights sourced from 4 different DL frameworks' repositories: \textit{Keras}~\cite{chollet2015keras}, \textit{PyTorch}~\cite{pytorch},
\textit{TensorFlow(TF)}~\cite{tensorflow2015-whitepaper}, and \textit{TFLite}~\cite{tensorflow2015-whitepaper}.
Each model is run through \texttt{DeltaNN} to generate model variants with different compiler optimization levels and \textit{target} DL frameworks, and executed on 4 GPU devices (discussed in Section~\ref{subsec:devices}).
In total, we evaluate a combination of 3 models, 12 DL framework conversions, 4 devices, and 3 optimization levels.


\subsection{Research Questions}
Our experiments are aimed at evaluating:
(1) Robustness of model output, by recording the top-1 output label for every combination of environment parameters and performing pairwise comparisons; and
(2) Robustness of model execution time, by measuring average inference time across executions in our dataset and comparing across different configurations. 
\noindent We investigate the following research questions for evaluating robustness:
\begin{description}[topsep = 0pt, itemsep = 0pt]
\item[Output Label Robustness]
\item[\quad RQ1. Label Sensitivity to DL Framework Conversions]  
\textit{Are the output labels of an image recognition model affected when converting the model from a source to a target DL framework?}
Both source and target frameworks are one among PyTorch, TF, TFLite, or Keras with $\mathit{source} \neq \mathit{target}$.
All conversions are through the intermediate ONNX format.
We compare output labels of target against the source for each image to check if any errors were introduced by model conversion.
We do this for each of the three image recognition models -- MobileNetV2, ResNet101V2, and InceptionV3. 
\item[\quad RQ2. Label Sensitivity to Compiler Optimizations]
\textit{Are the output labels of an image recognition model affected when changing the level of compiler optimization?}
We vary the optimization level within TVM between Basic, Default, and Extended and observe if there are any difference in the output label for images in the dataset. 
\eject 
\item[Inference Time Robustness]
%
\item[\quad RQ3. Time Sensitivity to DL Framework Conversion]
\textit{Are the inference times of an image recognition model affected when converting the model from a source to a target DL framework?}
\item[\quad RQ4. Time Sensitivity to Compiler Optimizations]
\textit{Are the inference times of an image recognition model affected when changing the level of compiler optimization?}
We are aware that differences in inference time is to be expected to some extent when changing
compiler optimizations.
The goal here is to identify unexpected performance degradation and extent of change with the different
compiler optimization levels. 

\end{description}


\subsection{Devices}
\label{subsec:devices}

We used four different hardware devices, featuring high-end to low-end GPU accelerators: 
\begin{itemize}
    \item an Intel-based server featuring an Nvidia Tesla K40c (GK11BGL) GPU (\textit{Server}),
    \item a Nvidia AGX Xavier featuring an Nvidia Volta GPU (\textit{Xavier}),
    \item a Laptop featuring an Intel(R) GEN9 HD Graphics NEO (\textit{Local}),
    \item and a mobile-class Hikey 970 board featuring an Arm Mali-G72 GPU (\textit{Hikey}).
\end{itemize}
For all GPU devices, we generate OpenCL device code, except for the Xavier device where we generate CUDA code, since it does not support OpenCL.
We found no accuracy impact between the two programming paradigms, and OpenCL vs. CUDA trade-offs are already explored\cite{cudavsopencl}.
We run the test dataset through the model configurations, and take the average inference time. 


\subsection{Dataset}

We use the ImageNet object detection test dataset~\cite{ILSVRC17} in our experiments, consisting of 5500 RGB images that are generally resized to 224$\times$224 pixels, and perform classification of 1000 possible labels and measure inference time on each image.
For models native to TensorFlow and TFLite, we observed that models actually used input size of 299, rather than the typical 244.
In general, using larger input sizes could increase the potential accuracy of the models, but also the computational requirements they need to perform.


\subsection{Execution Issues}
\label{subsec:execution_issues}

\begin{figure}
     \centering
     \includegraphics[width=0.99\linewidth]{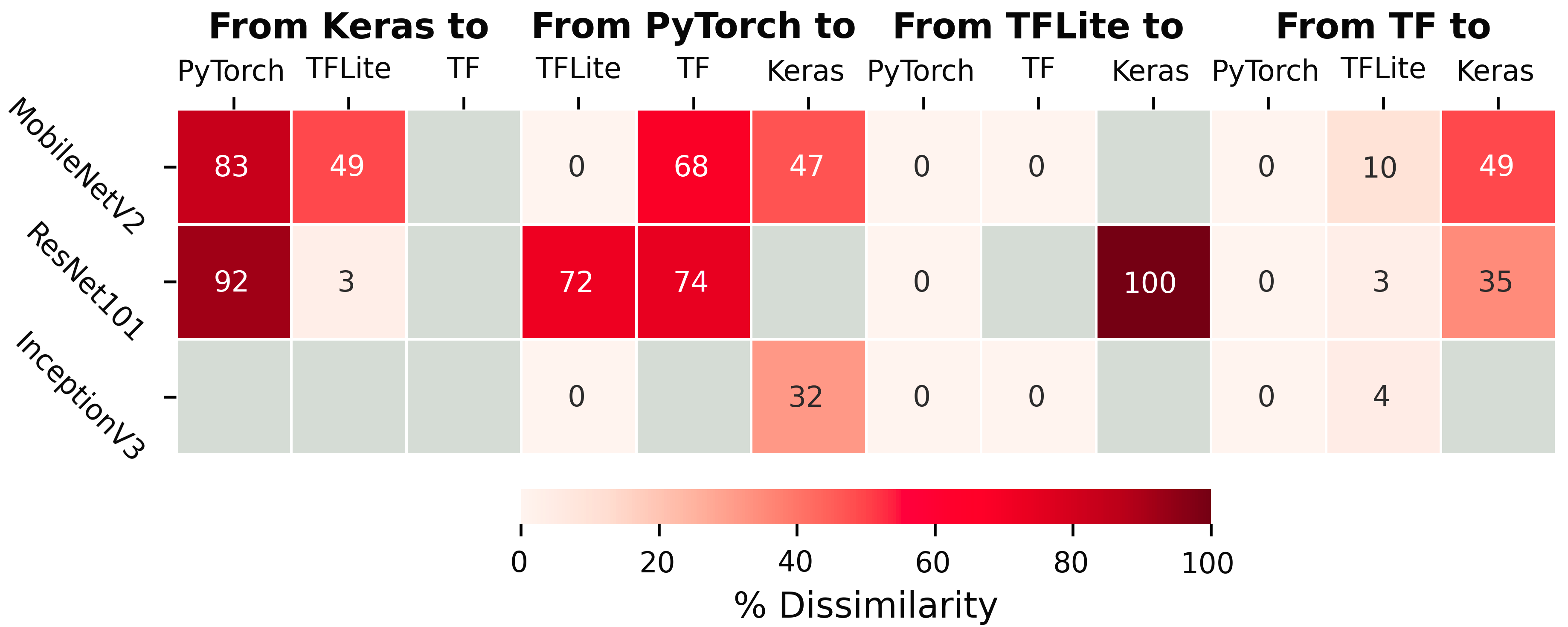}
     \caption{Pairwise comparison of output labels between \textit{source} and \textit{target} for a given model architecture across all images in the dataset.}
     \label{fig:conversionsorig}
\end{figure}

All environment parameter combinations could not be executed with all models due to the following incompatibility issues.
First, for ResNet101 sourced from PyTorch, we selected the V1 version the model instead of V2 as the V2 version was not provided in the official PyTorch repository. 
The version difference may have a larger effect on model inference time when we compare across DL frameworks.
Second, regarding MobileNetV2, we experienced problems when executing it on the Xavier device, as we received a \texttt{CUDA\_ERROR\_INVALID\_PTX} error, in all cases except when natively sourced from PyTorch.
Thus, we do not consider this device configuration for MobileNetV2 in our experiments. 
Third, while utilizing the conversion process, we encountered various cases where the source model conversion failed, as presented in Figure~\ref{fig:conversionsorig}. 
This happened due to incompatibility either of the source model with the conversion tool or the generated model operations in TVM.

\section{Results}

We present results with respect to discrepancies observed in output labels and inference time over the different DNN model configurations in the context of the research questions presented in Section~\ref{sec:experiments}.

\subsection{Robustness of Output Label Prediction}

\subsubsection{RQ1: Label Sensitivity to DL Framework Conversions}
\label{subsec:dlframeworkconversions}
The results are presented in Figure~\ref{fig:conversionsorig}, showing the degree of dissimilarity between source and target models.
As can be seen from the empty gray boxes, the conversion tool crashes in 11 out of the 36 conversions across the three DNN models, indicating that the conversion process failed. 
This happened due to compatibility issues between the conversion tool and a given model architecture, or the source or target DL framework. 
For instance, we could not convert any Keras models to TensorFlow, due to the \texttt{tf2onnx} tool being unable to handle some tensor element values. 
Additionally, we observe further 15 cases where the conversion succeed without crashing, but the target model gave different labels from the source model, with 11 of these cases having a significant discrepancy (over 35\%).
In particular, we observe a 100\% discrepancy in the output labels when converting the ResNet101 model from TFLite to Keras.

For conversions between either TF or TFLite and PyTorch, we observe no errors introduced by the conversion process across all models, while when converting TF to TFLite, we see relatively small discrepancies, 3-10\%, demonstrating more reliable conversion.
For TFLite to TF we had no discrepancies, but had one conversion failure (ResNet101).
This relative success is reasonable to expect, since TFLite has overlap with the TensorFlow codebase.
However, ideally the differences should all be 0\%.
Table~\ref{tab:native_accuracy} shows that the native accuracy of the TensorFlow and TFLite models are all identical, implying that
(1) the models are the same, and thus 
(2) the TFLite authors had 100\% success with their conversions.
However, we observe divergences using common open source conversion tools with default configurations.

Finally, the conversion of TF models to Keras gives varying results across models, with MobileNetV2 having 49\% dissimilarity, ResNet101 having 35\%, and InceptionV3 giving a model crash.
This points to weaknesses in the conversion tool with certain DNN model architectures. 

\begin{figure}
 \centering
 \includegraphics[width=0.99\columnwidth]{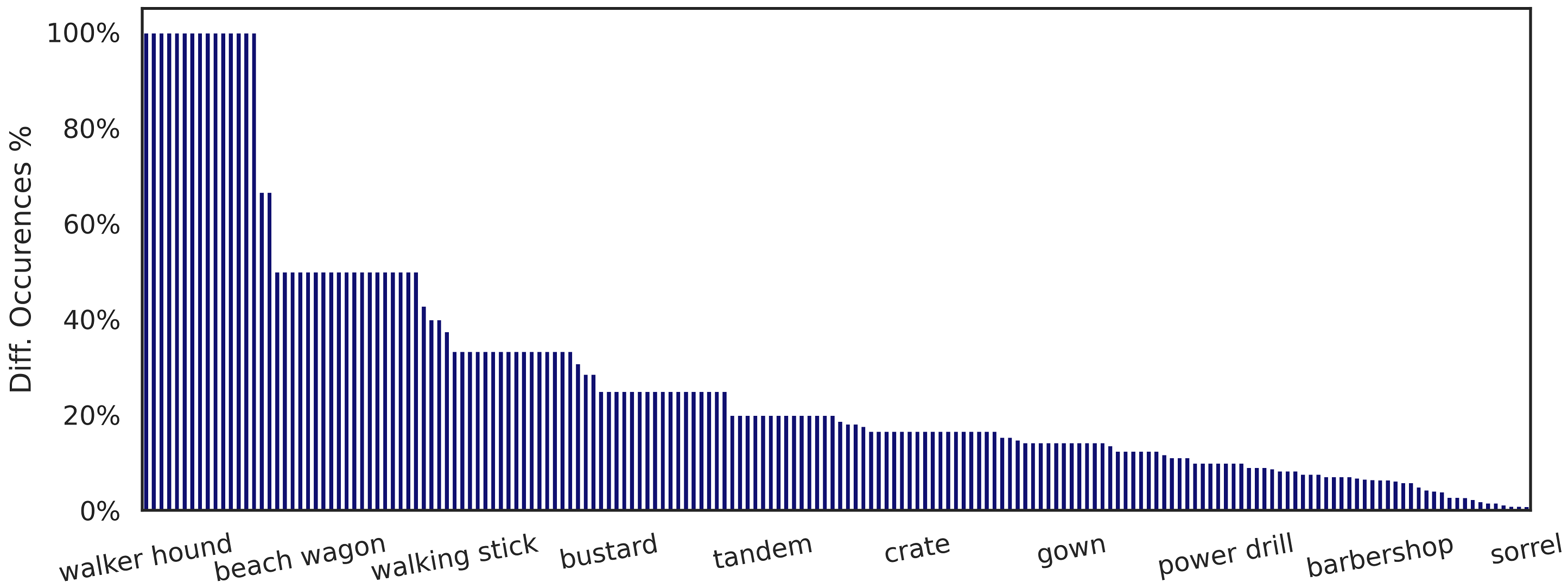}
 \caption{Percentage of affected images due to library conversions, InceptionV3, TF-to-TFLite conversion.}
 \label{fig:affected_labels}
\end{figure}

\paragraph{Fault Analysis}
We use the \textbf{Analysis} part of \texttt{DeltaNN} to explore in greater detail the cause of discrepancies across DNN model conversions. 
To illustrate the analysis in depth, we select one of the models and conversions that results in a discrepancy, InceptionV3 with TF as the source DL framework converted to target TFLite.
The discrepancies observed across \textit{source-target} is 4\%, and in Figure~\ref{fig:affected_labels} we show the class breakdown of the images which demonstrated differences, sorted by what proportion of that class showed discrepancies.
We highlight a subset of the class labels on the x-axis.
We observe that some classes are impacted more than others, with some classes such as ``walker hound'' disagreeing on 100\% of the images. However, this graph is not indicating test set accuracy, instead it is about agreement between the \textit{source} and converted \textit{target} models, which under ideal circumstances we would expect to have equal agreement in all cases.

We also performed inference using the intermediate ONNX format, which is used as part of the conversion and model loading process to TVM.
We used \textit{ONNXRuntime}~\cite{onnxruntime} for that purpose.
For all selected images, we found that the TF model differed from the TFLite inference results, whereas the TFLite model results were identical to ONNX, as seen in Table~\ref{tab:optcomparisonsconversions} for five images as an example.
The ground truth for these images matches the results from the source model, and deviates from TFLite and ONNX.
This narrows down the source of the error mainly to the conversion tool from the source TF model to TFLite, i.e., \texttt{TFLiteConverter}~\cite{tfliteconvertmodels} of the native TFLite API, and less to \texttt{tf2onnx}~\cite{tf2onnx}, which is widely used in the community (1.8k stars on GitHub).
We store the tensor outputs from the source and target models for further analysis.

\begin{table}[!ht]
\centering
    \caption{Inference (top-1 prediction) of 5 ImageNet images posing different results between models using TF, TFLite converted from TF, and complementary ONNX applied on InceptionV3 and run on Local device using Default optimization.}
    \label{tab:optcomparisonsconversions}
    \begin{tabular}{|l|l|l|l|}
    \hline
        Image ID & TF & TFLite (TF) & ONNX \\ \hline
        00001219 & scooter & moped & moped \\ \hline
        00002078 & cottontail & llama & llama \\ \hline
        00002439 & wallet & purse & purse \\ \hline
        00003928 & black grouse & bee & bee \\ \hline
        00004898 & wallaby & It. greyhound & It. greyhound \\ \hline
    \end{tabular}
\end{table}

\begin{figure} [t]
 \centering
\includegraphics[width=0.99\columnwidth]{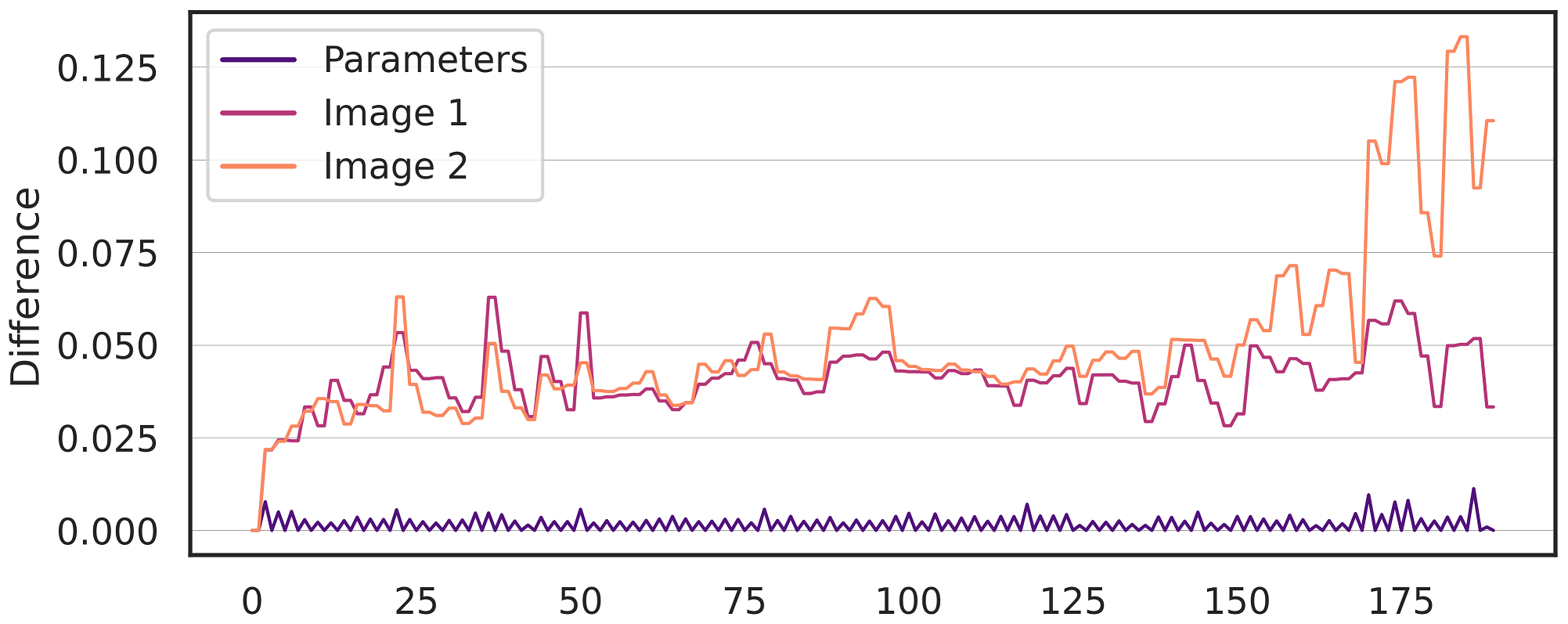}
 \caption{\label{fig:per_layer_differences} Layer-wise evaluation of the differences between a model sourced from TensorFlow, and converted to TFLite. 
 ``Parameters'' shows the mean difference between their weights and biases. \textit{`Image 1'} and \textit{`Image 2'} show models' differences in activations for two inputs.
}
\label{fig:faultlocalizationanalysis}
\end{figure}

Next, we performed execution on the \textit{source} and the \textit{target} model utilizing \texttt{DeltaNN}'s Debug execution, which relies on TVM's debugger and provides metadata about the execution.
Following this process, we perform per-layer activation analysis combining the debugger metadata with metadata of the build process for the \textit{source} and the \textit{target} models.
We compare the average differences between the models across layers utilizing parameters (i.e., weights and biases from the convolution layers), per-layer tensor outputs (i.e., activations), as well as the hyperparameter values for the respective layers.
We illustrate this for two images in Figure~\ref{fig:per_layer_differences}, focusing on the convolutional layers, where \textit{Image 1} generated the same output label across source and target models, whereas \textit{Image 2} produced completely different labels across source and target models for the top-5 predictions. 
We observe that both images have divergences in their activations, but for \textit{Image 2} the divergences are higher for later layers (layer 170 onwards).


\paragraph{Per Layer Activation and Model Parameter Analysis}
Figure~\ref{fig:per_layer_differences} highlights the difference between intermediate activation maps (i.e., the outputs of individual layers during execution), as well as the differences in the parameters.
We would expect the models to behave the same, since they should be the same model architecture and parameters.
Our observation is however that the output labels are not always consistent.
For \textit{Image 1}, both \textit{source} and \textit{target} versions of the model produce the correct label, even though their intermediate activations are between $0.0$ and $0.06$ on average.
However, for \textit{Image 2} the models disagree on the output label, and for later layers, we observe a higher average difference, up to around $0.13$.
This may imply that whatever error has occurred may have impacted later layers more, but this does not explain why this is not the case for \textit{Image 1}.

To identify the source of these discrepancies, we examine the differences in the parameters of the models (seen as the green line in Figure~\ref{fig:per_layer_differences}), which indicates a possible source of the error.
Across parameters, we observe a divergence of $0.0003$ on average and $0.011$ at most. 
In principle, when we run our conversion tool the parameters should be unchanged, i.e., bit-wise identical from TF to TFLite.
Furthermore, we presume that the model parameters are incorrectly copied at some stage of the DL framework model conversion process.
With this bug identified and fixed in the conversion tool, we could expect that the difference between the models goes away.
We demonstrate this by replacing the parameters of the converted model with the source model within TVM, and observe that 100\% of our divergence disappears. 
However, we cannot assume that these bugs will always be fixed.
Thus, we could also mitigate the impact of the error during training by simulating the conversion tool noise into our parameters, so that the model learns to be robust against it.

Therefore, the confirmation of our hypothesis still does not explain why we observed non-uniform divergence in the activation maps and output labels; despite the fact that the parameters had small, relatively uniform noise, and were identical between \textit{Image 1} and \textit{Image 2}.
However, if we reason about the underlying operations and mechanics of a DNN model, we can begin to make sense of it. 
We observe that the impact of these weight errors are cumulative, since weights are generally used repeatedly in multiply-accumulate operations.
Then, layers with more such operations (e.g., ReLU activation functions) will be more likely to have higher errors.

By complementary examination of the conversion tools involved, we identify that \texttt{tf2onnx} did not introduce any weight deviations, but that \texttt{TFLiteConverter} was responsible for introducing the fault.


\subsubsection{RQ2: Label Sensitivity to Compiler Optimizations}

We conducted experiments across all DL frameworks and device combinations described in Section~\ref{sec:experiments}, using only the native DNN model definition and varying the optimization level (Basic, Default and Extended), in order to observe inference time and output label discrepancies.

We found that varying compiler optimization levels causes no discrepancies in output labels for all three models. 
The lack of discrepancies/sensitivity is notable, since the Extended (\texttt{-o4}) level enables unsafe math optimizations that allow code violating IEEE \texttt{float} conventions to be generated. 
The conclusion is that these potential unsafe perturbations were small enough that all three models were resilient to them. 
It is however worth considering robustness checks with respect to optimization levels in safety-critical domains, in case that unsafe optimizations result in undesirable model outputs.
It is also not a foregone conclusion that the ostensibly semantic preserving optimizations of Basic and Default optimization levels would have produced no label divergences, as bugs in compilers are a common occurrence.
However, TVM's optimizations do not introduce any errors in our experiments.


\subsection{Robustness of Model Inference Time}


\subsubsection{RQ3: Time Sensitivity to DL Frameworks}

\begin{figure}[!t]
 \centering
 \includegraphics[width=\columnwidth]{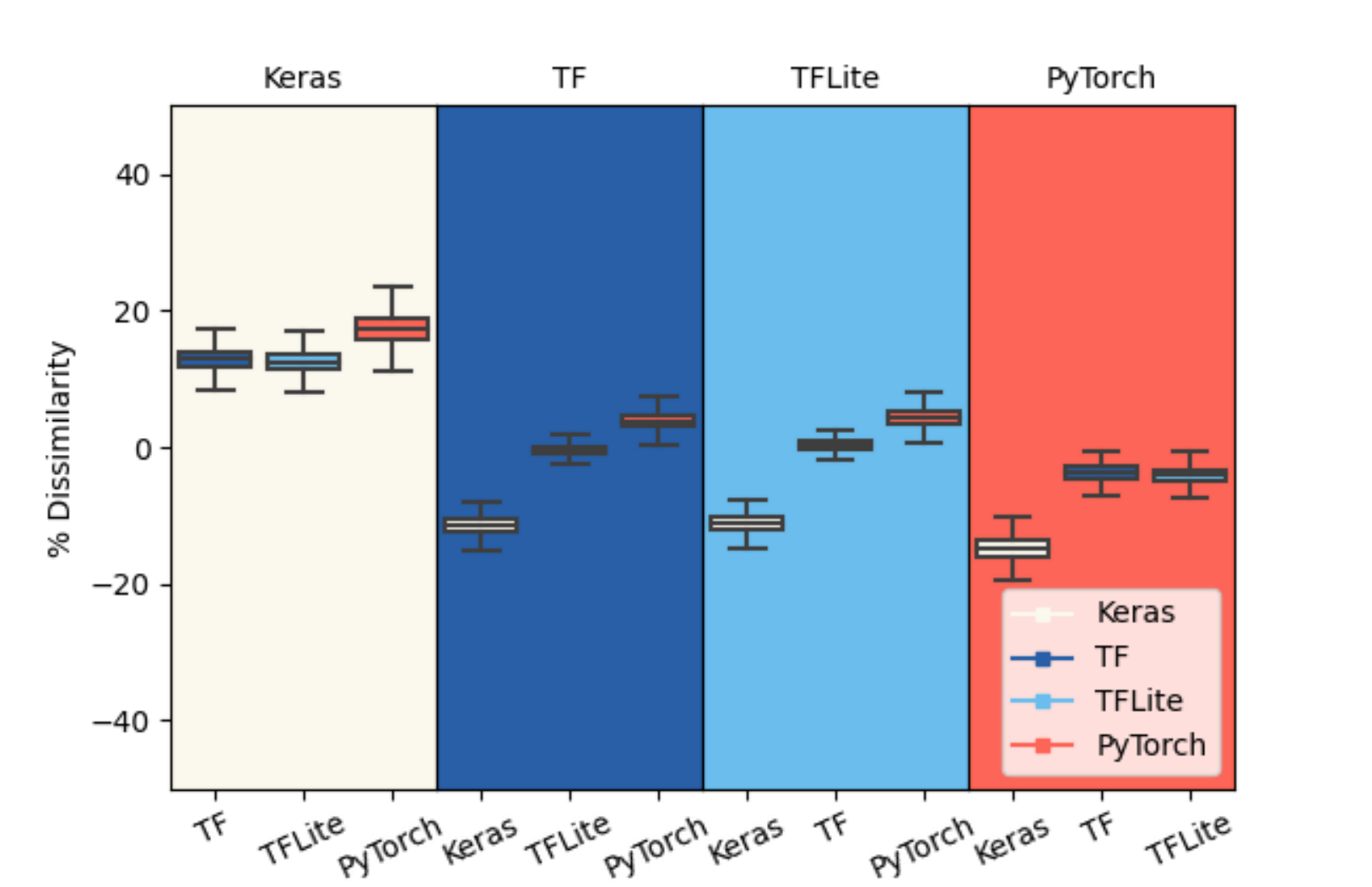}
 \caption{Inference time differences (\%) between DL frameworks on Server, for MobileNetV2, with Default Optimization.}
 \label{fig:exectimeframeworks}
\end{figure}

Between Keras and PyTorch native models, we observed a 4-16\% difference in inference time using the MobileNetV2 model with the Default optimization, deployed on the Server across models, with the largest difference of 16\% being seen between Keras and PyTorch, as shown in Figure~\ref{fig:exectimeframeworks}. 
The differences were confirmed to be significant using one-way ANOVA with 5\% significance level. 
We believe the difference is due to the different graph representation after framework conversion, e.g., one framework may represent a fully-connected layer as a ``dense'' operation and another may represent it as a ``batch matmul''; or some conversion tools may apply some of the graph-level optimizations such as batch-normalization fusion, so that even with a Basic optimization level the model is simplified.
We plan to investigate this further in future work. 

\subsubsection{RQ4: Time Sensitivity to Compiler Optimizations}

We observed a maximum speedup of 114\% in inference time with increasing optimization levels.
As part of our statistical analysis, we confirmed the observation to be significant using one-way ANOVA with 5\% significance level. 
This is not surprising, as different optimization settings have a direct effect on code efficiency.
Interestingly, there were instances where increased optimization led to a slowdown in inference time. 
For instance, MobileNetV2 from Keras and Extended optimization was 81\% slower than Basic on the Hikey device. 
We also confirmed our observations using one-way ANOVA.

To explore the impact of the compiler in greater detail, we enabled optimization passes individually, taking each optimization concept and applying it to the model separately, rather than using multiple optimizations together as the \texttt{-oX} bundles that we use at a high level.
We conducted an analysis on 100 images, using one optimization pass per-case, to understand which optimizations contributed to speedup or slowdown in model inference time for ResNet101 and InceptionV3 using TensorFlow and PyTorch DL frameworks\footnote{Our results can be found at: \url{https://github.com/luludak/deltann-results}}.

We found that no single optimization led to a significant inference time change for this experiment, suggesting that the non-trivial interactions between optimization passes are what contribute to these changes, making analysis and performance optimization more challenging. 
For ResNet101, the optimization which provides convolution operators' scale axis folding degraded the performance by 2.71\% on the Local device, while the combination of parallel operators had a positive impact of up to 2.82\% compared to using Basic optimizations alone.
With InceptionV3, constant folding had a positive impact of 4.9\%, while combining parallel operators \textit{degraded} the performance by up to 5.47\% compared to Basic optimizations alone.
The difference in effect of optimizations between InceptionV3 and ResNet101 is likely due to the difference in their model architecture and data flows.
We plan to analyze the reasons behind this in future work. 

\begin{figure}[!t]
 \centering
 \includegraphics[width=0.99\columnwidth]{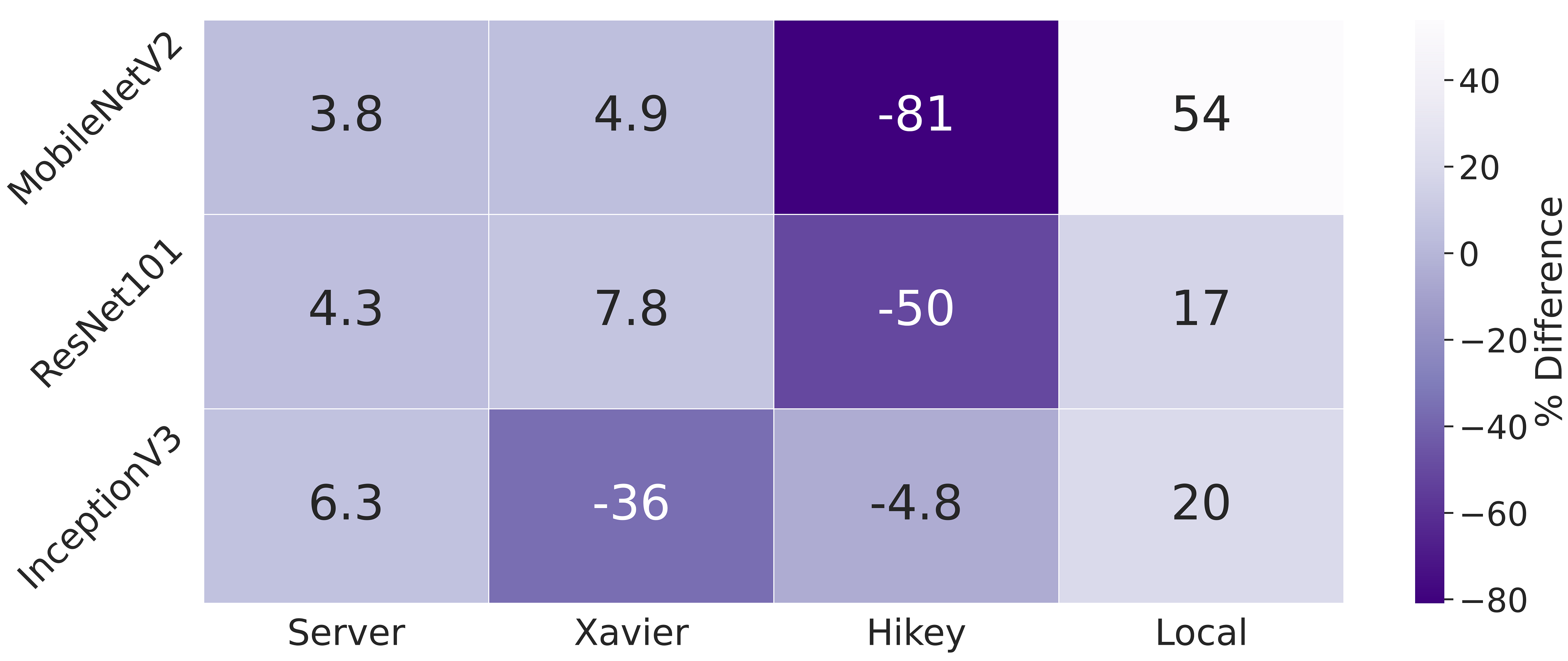}
 \caption{Inference time differences (\%) between Basic and Extended optimizations across devices, with native models from PyTorch.}
 \label{fig:exectimedifferences}
\end{figure}

However, we observed that the combination of optimization strategies can lead to significant performance degradation under certain contexts, such as low-end hardware acceleration devices. Figure~\ref{fig:exectimedifferences} shows the percentage difference in inference time for Basic versus Extended optimization on different devices and models with PyTorch, as an indicative example.
For each device, we find that times generally improve with increased optimization in the range of 3.8-8.4\% for Server, and 17-54\% for Local. 
Increased optimizations on Hikey, however, had a 81.8\% slowdown (confirmed with One-way ANOVA 5\%).
The Xavier device also had a 36\% slowdown (confirmed with One-way ANOVA 5\%) when increasing the optimization level from Basic to Extended on InceptionV3 model. 
For low to mid-range devices, Xavier and Hikey, we experienced a slowdown with increased optimization, and believe that the limited GPU memory poses a problem for the optimizations with parallel operations in the Extended optimization setting, leading to additional wait times, context switches, and GPU data transfer time, which result in a slowdown.
Investigation of cache behavior, data transfer times between the CPU and GPU, and processor idle times to clearly identify the reasons for slowdown is subject for future work.


\subsection{Threats To Validity}

There are five main threats to validity in our experiments: 
\begin{enumerate}
    \item We only evaluate robustness using three image recognition models that are widely used. The results are model dependent as seen in our experiments and will likely vary on other models;
    \item We use the ImageNet~\cite{ILSVRC17} object detection test dataset for our experiments, which we believe adequately stresses configurations. 
    Other datasets may yield different robustness results on the models considered;
    \item Model pre-processing is crucial for model performance~\cite{preprocessingimportance}, and models may give suboptimal performance if given data with ineffective and erroneous pre-processing. 
    We use the recommended pre-processing for each model and DL framework from the official repositories extracted;
    \item Beyond the DL framework conversions explored in our results, we also have a ``hidden'' conversion step, i.e., importing models into Apache TVM, which itself may introduce errors.
    To ensure that errors are not introduced before loading each model into TVM, we generate ``target outputs'' from their source framework using an indicative number of random image samples.
    After importing into TVM, we confirmed that we match the target outputs, however this may not guarantee that the import process is entirely bug-free;
    \item We consider the potential deviations of inference time measurement. 
    To ensure that time deviations are taken into account, we repeat inferences 10 times for each image and use the average inference time across each run across a small-scale test dataset, verifying that no deviations happen on scaling.
    Note that non-trivial medium-term cache behavior may cause the inference time to change over time with repeated inferences, and depending on the deployment scenario of interest, only the ``first'' inference time may be of interest, or the inference time of the `$N$th' sample where $N$ is a large value.
\end{enumerate}

\section{Lessons Learned}
\setlist[description]{style=unboxed}
Our empirical study exploring the effect of changing computational environment parameters revealed the following findings:

\paragraph{Failures in DL Framework Conversion} 
Automated conversion of models between DL frameworks can introduce significant output label discrepancies. 
We observed up to 83\% output label dissimilarities when converting PyTorch to Keras for MobileNetV2.
In addition, converting Keras to TF generated failures for all three models under test.
These errors can be introduced in model weights, parameters, graph and architecture representation during the conversion process. 
Our analysis revealed errors in model weights introduced by \texttt{TFLiteConverter} when converting from the source model (TF) to TFLite, which can be fixed by correctly copying over the source model weights. 

\paragraph{DL Framework Conversion - Impact on Model Inference Time}
Changing the DL framework used to generate the model can have a considerable effect on model inference time.
The extent of this impact depends on other environment parameters.
Inference time impact varied from 1-16\% with the largest impact (16\%) observed between Keras and PyTorch.

\paragraph{Performance Degradation from Compiler Optimizations}
Compiler optimizations are generally expected to improve the performance of a model. 
However, our findings indicate that for certain scenarios defined by the device, model and library, compiler optimizations can be detrimental to model inference time.
In our experiments, we observed this performance degradation to the greatest extent when applying Extended optimization to MobileNetV2 for the Hikey device, which resulted in 81\% performance degradation when compared to a lower optimization level using Basic.

Finally, it is important to note that in safety-critical applications, the consequences of the above sensitivities can be crucial.
Therefore, it is essential that framework, compiler, and hardware communities, along with the developers of DNN models are aware of these sources of error, and test their systems for robustness to computational environment changes.
Currently, there is no regulation or benchmarking of DNN model performance and accuracy for environment parameter configurations. 
The results from our study indicate that assessing sensitivity to environment parameters is an important consideration during model development and use.

\section{Conclusion}
We introduced the \texttt{DeltaNN} Differential testing framework to explore the impact of computational environment parameters on image recognition models.
In particular, we study the effect of converting between popular deep learning frameworks (TensorFlow, Keras, TFLite, PyTorch), compiler optimization settings, and hardware devices on the output labels of three widely used image recognition models.
We also monitor the impact of these parameters on model inference time. Overall, we find that conversions between DL frameworks significantly impact output labels of the DNN models by up to 100\%. 
Our framework also provides analysis capabilities for label discrepancies stemming from framework conversions.

\bibliographystyle{IEEEtran}
\balance
\bibliography{00-main.bib}

\end{document}